\documentclass{INTERSPEECH2023}

\interspeechcameraready

\usepackage{subcaption}
\title{Efficient Spoken Language Recognition via Multilabel Classification}
\name{Oriol Nieto, Zeyu Jin, Franck Dernoncourt, Justin Salamon}
\address{
  Adobe Research, San Francisco, CA, USA}
\email{onieto@adobe.com}

\begin{document}

\maketitle
 
\begin{abstract}
Spoken language recognition (SLR) is the task of automatically identifying the language present in a speech signal. Existing SLR models are either too computationally expensive or too large to run effectively on devices with limited resources. For real-world deployment, a model should also gracefully handle unseen languages outside of the target language set, yet prior work has focused on closed-set classification where all input languages are known a-priori. In this paper we address these two limitations: we explore efficient model architectures for SLR based on convolutional networks, and propose a multilabel training strategy to handle non-target languages at inference time. Using the VoxLingua107 dataset, we show that our models obtain competitive results while being orders of magnitude smaller and faster than current state-of-the-art methods, and that our multilabel strategy is more robust to unseen non-target languages compared to multiclass classification.

\end{abstract}

\section{Introduction}
\label{sec:intro}


Automatic speech recognition (ASR) techniques have achieved near human accuracy for various languages~\cite{wang2019end, amodei2016deep}, and enable applications such as text-based audio/video editing~\cite{jin2017voco,fried2019text},  voice translation~\cite{Federico2020}, and virtual assistants~\cite{kepuska2018next}. 
When the language is not known in advance, Spoken Language Recognition (SLR) is often a necessary first step before running ASR, as most ASR systems require this information (the target language) to correctly transcribe the speech signal. 
Past research in SLR has been organized around challenges such as NIST Language Recognition Evaluations~\cite{sadjadi20222021}, focusing on improving accuracy with large neural networks and extensive data. Recent advances in large-scale and self-supervised models have achieved impressive generalization across hundreds of languages with near-perfect accuracy~\cite{Radford_undated-xg, Babu2021-co, efficient2022}. 

Despite the significant progress, there are two important limitations that make real-world deployment of such models challenging in resource-constrained scenarios: (1) models are too large or resource intensive for devices with limited compute power, (2) they operate on a closed set of target languages, i.e., they are not designed to handle the scenario where an unknown language is presented to the model at inference time.
Most modern SLR systems are based on deep architectures, all of which contain millions if not tens or hundreds of millions of parameters.
One of the smallest models is based on X-vectors~\cite{Snyder2018-bh}, which were initially introduced for the task of speaker recognition~\cite{Snyder2018-tk}.
Such vectors are computed using blocks of Time-Delayed Neural Networks (TDNNs)~\cite{Peddinti2015-zb} and an aggregation layer, allowing for a large receptive field with small kernels, yielding good performance with a reduced amount of parameters. 
A larger and more powerful version of this model was recently introduced for the same task of speaker recognition, and it uses higher capacity TDNNs with an emphasized channel attention, propagation, and aggregation (ECAPA) set of mechanisms~\cite{Desplanques2020-hg}.
Such a model was later used successfully for SLR~\cite{Ravanelli2021-kj}.
ECAPA-TDNNs have around four times the number of parameters as the original X-vector model.
More recent and larger systems are capable of jointly recognizing a given language and solving additional tasks in a single pass.
For example, Whisper~\cite{Radford_undated-xg} can perform multi-lingual speech to text transcription, having an implicit SLR system embedded in the model.
This architecture is an encoder-decoder Transformer~\cite{Vaswani2017-fm} trained on 680k hours of speech.
The smallest version of this model has almost 2x the parameters of ECAPA-TDNN, and its medium version is over an order of magnitude larger than its smallest counterpart.
Finally, XLS-R~\cite{Babu2021-co} is another self-supervised model whose smallest version has around 300 million parameters.
It is based on Wav2Vec 2.0~\cite{Baevski2020-sn} and is pre-trained with over half a million hours of speech covering 128 languages. 
When fine-tuned on the SLR task, it achieves state-of-the-art results on most test datasets. 
While self-supervised models lead the charts in terms of accuracy, they require significant compute power to operate.

There is a growing need for robust SLR models that can run efficiently embedded on-device. 
SLR on-device eliminates computation and networking costs involved with running SLR in the cloud, and can prevent tracking and other potential threats to user privacy.
However, large-scale models such as those discussed above are impractical to use on, e.g., a mobile device, due to size and runtime constraints. 
There are use cases where a user only requires on-device SLR for a limited number of languages (e.g., the languages they speak), not hundreds, providing an opportunity to trade-off the language set size for model efficiency.
Furthermore, the models in the studies discussed above are incapable of detecting unsupported languages: they assume the input is always one of the target languages, and would incorrectly classify a non-target language as one of the target languages the model was trained on. For real-world deployment, a model should be able to identify this scenario and ``fail gracefully'' by classifying the input as ``unknown'' or ``other''.

In this paper we address these two aforementioned limitations.
We investigate small-footprint models for SLR that can run effectively on-device. 
We propose novel variants of widely-used architectures, including TC-ResNets~\cite{Choi2019-ff} and ECAPA-TDNNs~\cite{Peddinti2015-zb}, and compare them to top-performing large models using VoxLingua107~\cite{Valk2021-uc}, a sizeable speech dataset recently introduced for SLR research~\cite{Babu2021-co}.
We show that our proposed lightweight architectures achieve competitive error rates with models that are two orders of magnitude larger in terms of parameters.
To handle non-target (unseen) languages, we propose a multilabel classification approach that is novel in the context of SLR, and show that it produces models that are more robust compared to modeling all non-target languages via a single ``Other'' class in the commonly used multiclass setup.
To the best of our knowledge, this is the first work to study SLR from an efficiency viewpoint and address the problem of non-target languages at inference time.

\section{Models for Efficient SLR}
\label{sec:models}

We explore two families of architectures for efficient SLR: Temporal Convolution ResNets (TC-ResNets)~\cite{Choi2019-ff}, and ECAPA-TDNN~\cite{Ravanelli2021-kj}. For the latter, we propose a modification that makes it significantly lighter, which we call LECAPAT.
For all the proposed models, our input is a log-melspectrogram with 64 mel-frequency bins computed from an input audio signal sampled at 16 kHz. The mel-spectrogram is computed using a 25 ms Hann window, FFT size of 64 ms, and a 10 ms hop size.

\subsection{Temporal Convolution Residual Networks}

\begin{figure}[tb]

\begin{minipage}[b]{.5\linewidth}
  \centering
  \centerline{\includegraphics[width=2.5cm]{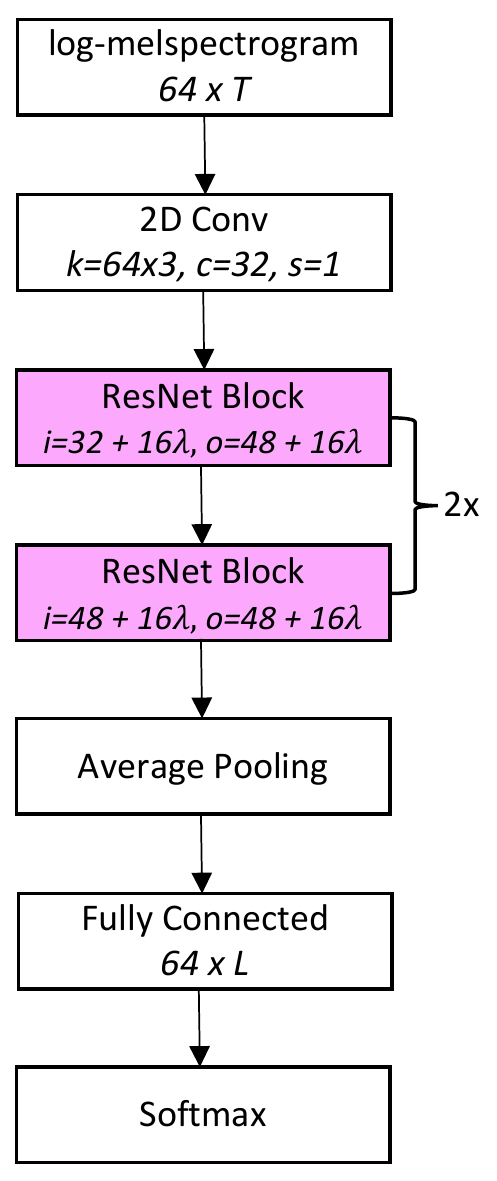}}
  \centerline{(a) TC-ResNet10}\medskip
\end{minipage}
\begin{minipage}[b]{.4\linewidth}
  \centering
  \centerline{\includegraphics[width=4cm]{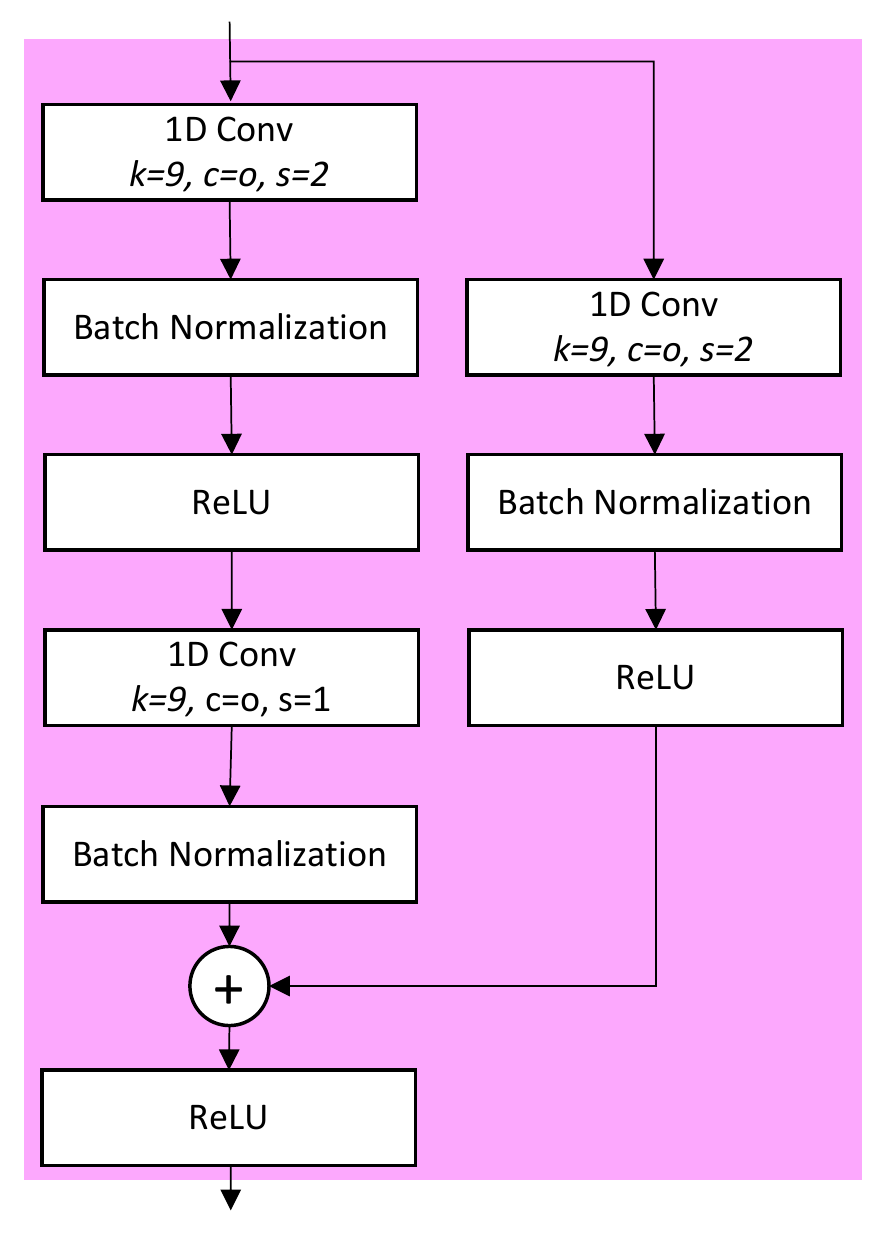}}
  \centerline{(b) ResNet Block}\medskip
\end{minipage}

\caption{Proposed TC-ResNet10 architecture, where $T$ is time frames, $c$ number of channels, $s$ stride, $L$ number of languages, and $i$ and $o$ input and output to the ResNet Block, respectively.}
\vspace{-0.5cm}
\label{fig:tcresnet10}
\end{figure}

Temporal convolution networks apply a 2D convolution to the input spectrogram with a kernel whose height matches the number of frequency bins,
compacting the frequency dimension into a set of 1D time representations (one per channel). Then 1D convolutions can be applied in the subsequent layers of the architecture, making it highly efficient in both number of parameters and inference time.
These networks were employed for early approaches to music recommendation using deep learning~\cite{NIPS2013_b3ba8f1b}, and more recently added residual connections~\cite{He2015-gk} for efficient keyword spotting~\cite{Choi2019-ff}.

We propose two different flavors of TC-ResNets for efficient SLR:
TC-ResNet10 is a slightly larger version of the model introduced in~\cite{Choi2019-ff}, but with fewer layers (10 as opposed to 14), depicted in Figure~\ref{fig:tcresnet10}. 
The two ResNet blocks are repeated (i.e., four ResNet blocks in total), where their size changes based on the repetition number $\lambda \in \{0, 1\}$. This model has 200k parameters.
The second model is TC-ResNet14, introduced by Choi et al.~\cite{Choi2019-ff}, with 100k parameters.

\subsection{LECAPAT: Light ECAPA Time-Delayed NNs}

\begin{figure}[tb]

\begin{minipage}[b]{.5\linewidth}
  \centering
  \centerline{\includegraphics[width=2cm]{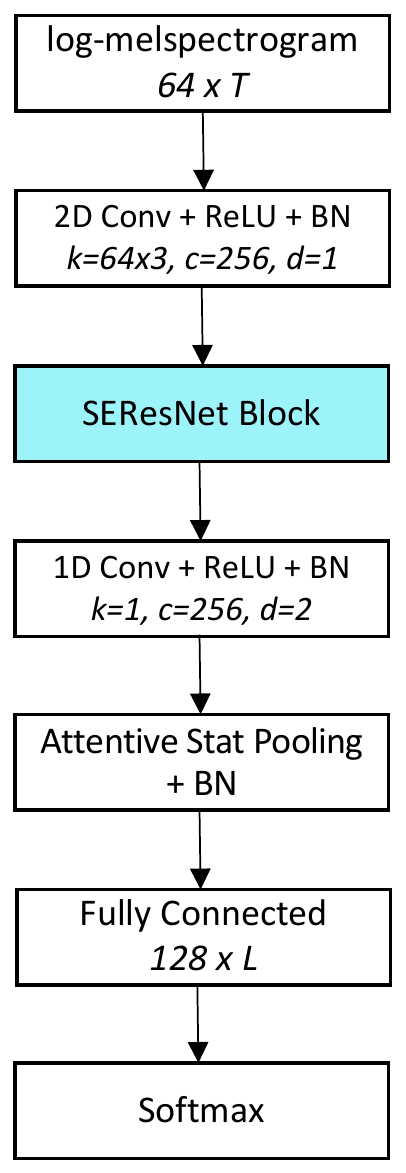}}
  \centerline{(a) LECAPAT}\medskip
\end{minipage}
\begin{minipage}[b]{.4\linewidth}
  \centering
  \centerline{\includegraphics[width=3cm]{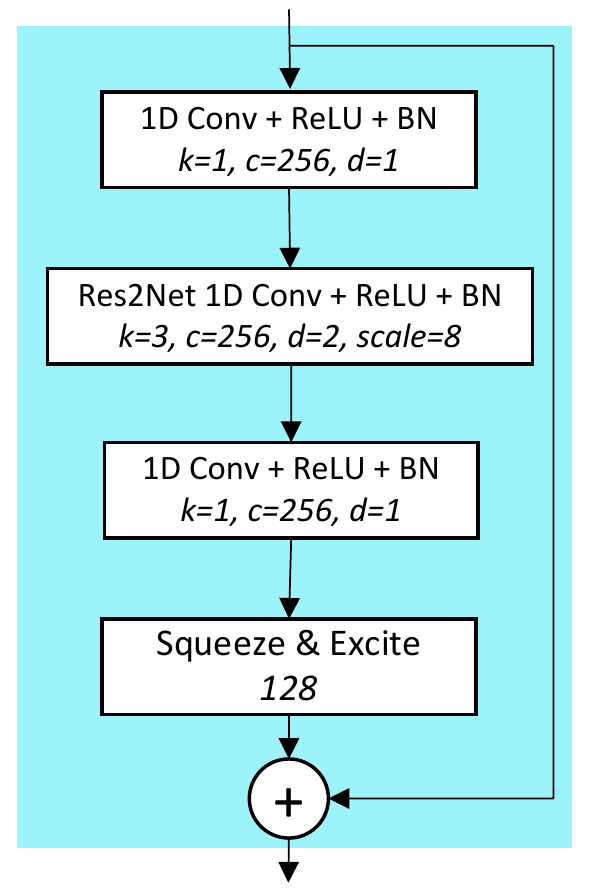}}
  \centerline{(b) SEResNet Block}\medskip
\end{minipage}
\vspace{-0.1cm}
\caption{Proposed LECAPAT architecture. $d$ is dilation size, BN is Batch Normalization~\cite{Ioffe2015-nr}, and \emph{scale} is the scale for the Res2Net model~\cite{Gao2019-yt}. All convolution strides are set to $s=1$.}
\vspace{-0.5cm}
\label{fig:lecapat}
\end{figure}

TDNNs~\cite{Peddinti2015-zb} were introduced to efficiently model long temporal contexts.
They use dilated convolutions to increase the receptive field, and are thus capable of modeling long-term structures without adding extra complexity in terms of computation or model size.
%
ECAPA-TDNNs~\cite{Desplanques2020-hg} added the following set of mechanisms: (i) per-channel attention, (ii) squeeze-excitation (SE) residual blocks that contain an SE mechanism~\cite{hu2018squeeze} and a Res2Net model~\cite{Gao2019-yt}, and (iii) an aggregation technique that attends to several levels of feature statistics across the architecture.
While more computationally expensive, ECAPA-TDNNs achieve better performance on speaker recognition, and have been shown to be highly effective for SLR~\cite{Ravanelli2021-kj}.

We propose a more efficient and lightweight version of ECAPA-TDNN, LECAPAT (for Light ECAPA-TDNN).
Our main intuition is that, while the original model was designed to recognize thousands or millions of speakers, SLR typically involves recognizing dozens or at most hundreds of languages, suggesting a model with reduced capacity may still perform well on this task.
We reduce the complexity of ECAPA-TDNN making it two orders of magnitude smaller: from 21 million parameters down to 600k.
LECAPAT employs a single SE residual block instead of three, 
and we reduce the number of parameters in each of the main blocks, as depicted in Figure~\ref{fig:lecapat}.

\subsection{Multilabel Classification}
\label{sub:multilabelclass}

ASR applications typically only support a subset of languages. Thus, an SLR model should recognize when a non-target language, i.e., a language that is not one of the languages the model is trained to recognize, is provided as input, to avoid sending it to be transcribed by ASR.
A common approach in multiclass tasks for handling non-target classes is to add a single ``Other'' class.
Our early experiments revealed that this approach, i.e., adding a single class for all non-target languages, resulted in poor performance for SLR.
The ``Other'' class must capture many different languages, including some that may be similar to a target language. Our conjecture is that the model struggles to simultaneously group all non-target languages \emph{and} separate them from the target languages. 
For example, in our experiments Spanish and German are target languages, while Catalan and Yiddish are not. The multiclass setup has to group Catalan and Yiddish and at the same time separate Catalan from the relatively close Spanish and Yiddish from the relatively close German, a challenging proposition.

Instead, we propose to train a \emph{multilabel} model for SLR, a novel approach in this context.
Unlike the multiclass setup which is forced to always return a single positive class, 
multilabel models allow for (1) several classes to be positive at the same time and (2) no positive classes for a given input.
Thus, the model can focus on representing the target languages and separating them from non-target languages, without explicitly modeling all possible non-target languages. 
Implementation-wise, we swap the final softmax layer in our models with sigmoids, such that zero positives are allowed.
We return the language with the highest output activation as the model's prediction, unless all activations are below a threshold (0.5) in which case the ``Other'' class is detected.
Note that this adds no extra complexity to the models during inference.

\section{Experiments}
\label{sec:experiments}

\subsection{Data}

We use the VoxLingua107 dataset~\cite{Valk2021-uc} for all of our experiments. The dataset consists of a training set and a ``development'' set for model evaluation.
The training set comprises 6,628 hours of speech from 107 languages, with an average of 62 hours per language, though it is heavily imbalanced. 
The development set contains 1,609 manually validated speech segments from 33 languages.

\begin{figure}[!t]
  \centering
  \centerline{\includegraphics[width=0.85\linewidth]{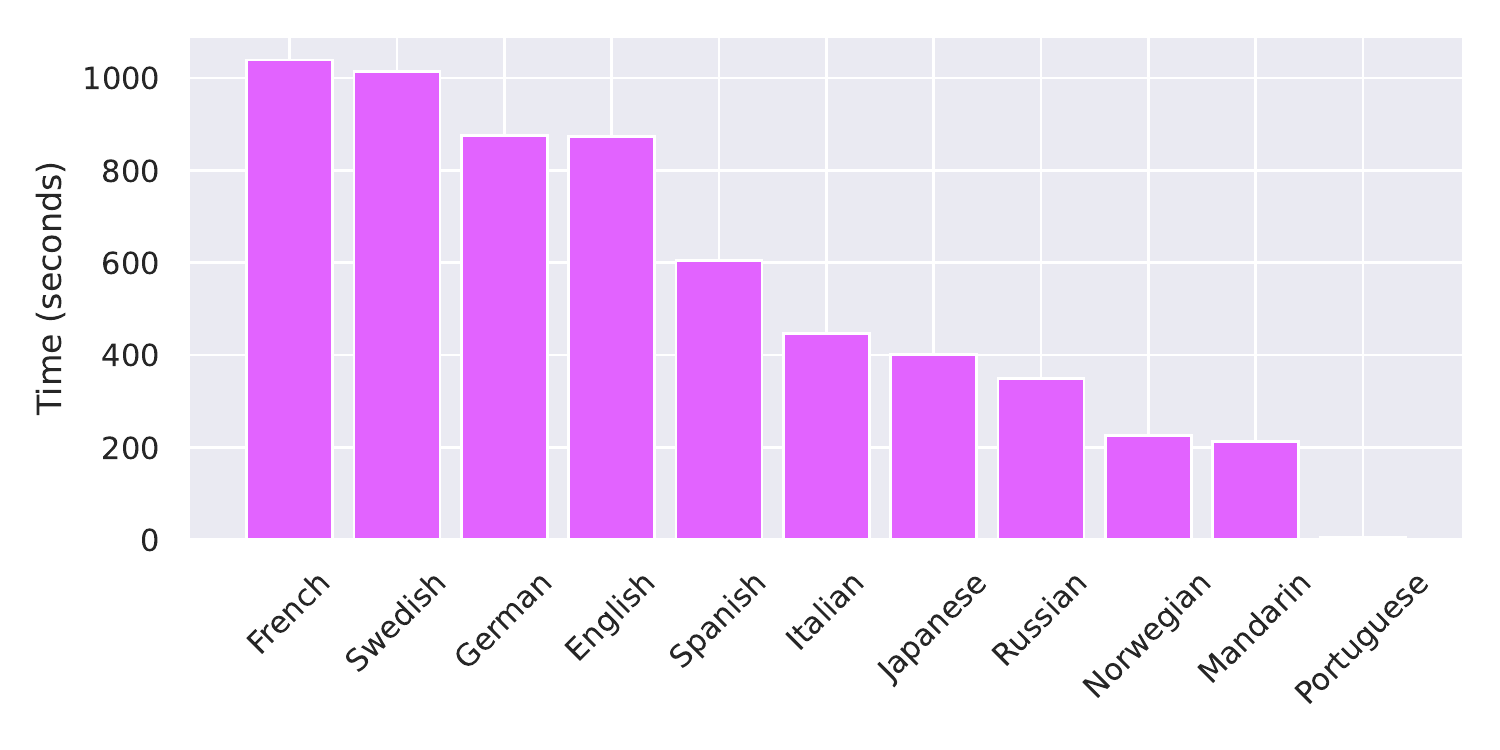}}
\vspace{-0.3cm}
\caption{Seconds per language in the VoxLingua11 test set.}
\vspace{-0.5cm}
\label{fig:voxlingua107s}
\end{figure}

Our focus is on real-world scenarios where the target language set is limited, as is often the case for speech applications, for example due to localization constraints. This common scenario is one for which efficient models are particularly promising as, we hypothesize, the reduced language set suggests we should be able to train performant models with significantly lower capacity compared to models targeting hundreds of languages.
To this end, we construct a new test set taking a subset of $L=11$ languages from the VoxLingua107 development (evaluation) set, which are the languages for which speech-to-text (ASR) is supported in a commercial video editing tool. We call this test set VoxLingua11, with the distribution of seconds per language depicted in Figure~\ref{fig:voxlingua107s}. We maintain the class imbalance for these languages from the VoxLingua107 development set, to make VoxLingua11 easy to reproduce.

We use VoxLingua11 to evaluate performance of our models and baselines under the multiclass setup where only target languages are included in the test set.
Since our final goal is to evaluate our models when non-target languages are present in the test set, we also create VoxLingua11+O, which expands VoxLingua11 with an ``Other'' class containing 128 samples with at least one sample from each of the non-target languages in the VoxLingua107 development set. VoxLingua11+O can be used to evaluate models under both the multiclass and multilabel setups, where in the former models have 12 output neurons (11 target languages + ``Other''), and in the latter they have 11 output neurons and ``Other'' is predicted as described in Section \ref{sub:multilabelclass}.

To make the models robust to real-world recordings, we augment the training set with noise, reverb, and random equalization following the work of Su et~al.~\cite{Su:2021:HSS}.
Since the training set is imbalanced, 
we use the augmentation to balance our training data, such that each training epoch contains a balanced sampling of the target languages.
The ``Other'' class is treated as a 12th language by randomly sampling across the non-target languages during training.
%
All the models evaluated in this work take log-melspectrograms of 10-second audio clips as input.
If the input recording is shorter than 10 s, we center it and zero pad to the left and right to obtain a 10 s clip.
If it is longer, we take a random 10 s subclip during training. At test time, for longer recordings we slide a 10 s window over the input with a 5 s hop size and average the predictions over time.

\subsection{Baselines}

We use the SLR models reviewed in Section~\ref{sec:intro} as baselines.
The open source implementations of the X-Vector\footnote{\scriptsize{\url{https://github.com/KrishnaDN/x-vector-pytorch}}} and ECAPA-TDNN~\cite{speechbrain} are trained from scratch on the VoxLingua107 training set.
We verify that we obtain similar results to those reported in the original publications.
For the larger baselines, XLS-R and Whisper, we rely on their published weights, as these higher capacity models were trained on a diverse conglomerate of datasets--including VoxLingua107--that make up much larger amounts of data.
We use the 300M parameter (i.e., smallest) version of XLS-R fine-tuned on VoxLingua107\footnote{\scriptsize{\url{https://huggingface.co/TalTechNLP/voxlingua107-xls-r-300m-wav2vec}}} and the ``Medium" and ``Tiny" versions of Whisper~\cite{Radford_undated-xg}.

\subsection{Metrics and Optimization}

We report the models' error rate $err$, the standard for SLR evaluation, where $err=100(1-acc)$ and $acc$ is the multiclass classification accuracy.
We report the inference runtime of each model on VoxLingua11 using either a CPU (Intel Cascade Lake on a g4dn.4xlarge G4 AWS instance) or a V100 GPU as a real-time factor (rtf) coefficient, i.e., the total test set audio duration divided by the total model inference time (larger rtf = faster).

Our proposed models are trained on the augmented VoxLingua107 training set using a V100 GPU, the Adam optimizer ($\beta_1{=}0.9$ and $\beta_2{=}0.999$), early stopping, and hyperparameter Bayesian optimization to determine learning rates and batch sizes, which fluctuate between $[10^{-3}, 10^{-5}]$ and $[32, 128]$, respectively.
We minimize the categorical and binary cross-entropy loss for multiclass and multilabel models, respectively.

\section{Results}
\label{sec:results}

\subsection{Accuracy versus size and runtime}
\label{sec:results_tradeoffs}

\begin{figure}[tb]
  \centering
  \centerline{\includegraphics[width=\linewidth]{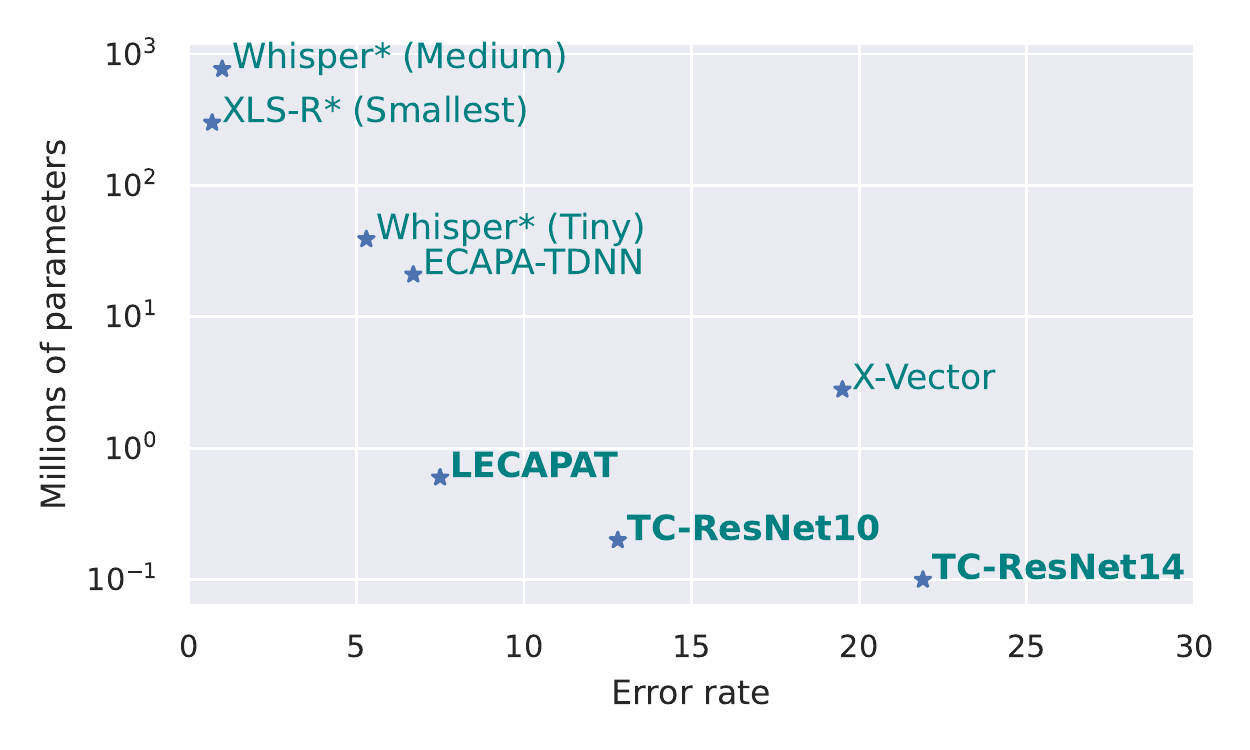}}
\vspace{-0.3cm}
\caption{
Model size versus error rate, multiclass classification on VoxLingua11. Proposed models in bold. * indicates original weights used, all other models were trained from scratch.
}
\label{fig:results}
\end{figure}

We start by examining the trade-off between model capacity and error rate, depicted in Figure \ref{fig:results}, with our proposed models in bold. Results are shown for multiclass classification on VoxLingua11 so that we can compare to all baselines. The models trained from scratch only see the 11 selected languages from VoxLingua107 during training. The error rates are also reported in Table \ref{tab:times}. Note the log scale of the parameters (y) axis. Expectedly, we see a correlation between model size and error rate in both the baselines and our proposed models. For the baselines, we see Whisper (Tiny) and ECAPA-TDNN provide an order-of-magnitude (OOM) reduction in model size compared to Whisper (Medium) and XLS-R (20-40M vs 300-800M) with a relatively low drop in error rate of 4-5 points. X-Vector is a further OOM smaller but its error rate is significant. 

Most interestingly, we see that our proposed LECAPAT model performs very closely to the Whisper (Tiny) and ECAPA-TDNN baselines with an error rate of 7.5, but is a whole two OOM smaller (and four OOM smaller than the largest baselines), with just 0.6M parameters. The TC-ResNet models, though smaller still, are notably less accurate, thus providing a less desirable tradeoff. LECAPAT (and the ResNets) are also significantly faster to run inference with, as shown in Table \ref{tab:times}, running over 800 times faster than real-time on GPU, which is 12 times faster than Whisper (Tiny) and 1.5 times faster than ECAPA-TDNN. It is interesting to note that despite their similar size and accuracy, the latter is around 8 times faster than the former, which shows that model size is not a good proxy for model runtime and highlights the importance of reporting runtime metrics when considering models for deployment. It is also noteworthy that LECAPAT remains fast on CPU, whereas ECAPA-TDNN becomes 6 times slower. This makes our proposed model a strong candidate for deployment scenarios where a GPU may not be available. While our best proposed model is outperformed by the strongest baselines in terms of accuracy, it trades off 6 accuracy points for a dramatic four OOM reduction in size and almost two OOM increase in speed.

\begin{table}[tb]
\begin{small}
\begin{center}
\begin{tabular}{ l c c c }
 Model & $err$ $\downarrow$ & rtf (CPU) $\uparrow$ & rtf (GPU) $\uparrow$ \\ 
 \hline
 XLS-R (Smallest) & 0.7 & 11.02 & 13.03 \\
 Whisper (Medium) & 1.0 &3.43 & 28.31 \\
 Whisper (Tiny) & 5.3 & 39.54 & 65.56 \\
 ECAPA-TDNN & 6.7 & 83.95 & 522.40 \\
 X-Vector & 19.5 & 524.2 & 656.54 \\ 
 \hline
 LECAPAT (Ours) & 7.5 &  678.42 & 810.93 \\
 TC-ResNet10 (Ours) & 12.8  & 966.91 & 821.56\\
 TC-ResNet14 (Ours) & 21.9 & 932.12 & 824.26 \\
\end{tabular}
\vspace{0.1cm}
\caption{Multiclass error rate and rtf on VoxLingua11.}
\label{tab:times}
\vspace{-0.6cm}
\end{center}
\end{small}
\end{table}

\subsection{Dealing with non-target languages}

\begin{figure}[t]
  \centering
  \begin{subfigure}{.23\textwidth}
  \centering
  \centerline{\includegraphics[width=\linewidth]{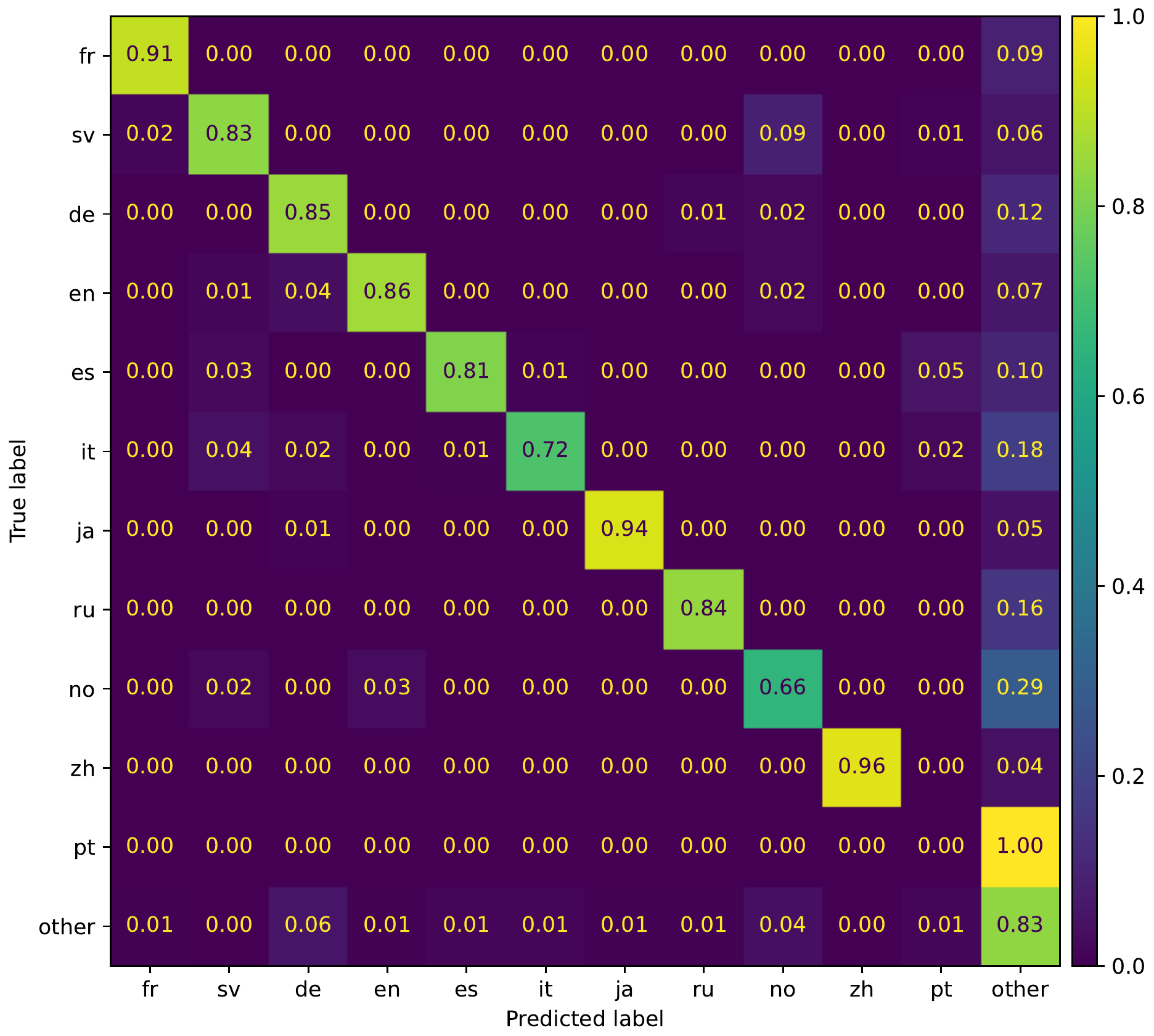}}
  \end{subfigure}
  \begin{subfigure}{.23\textwidth}
  \centering
  \centerline{\includegraphics[width=\linewidth]{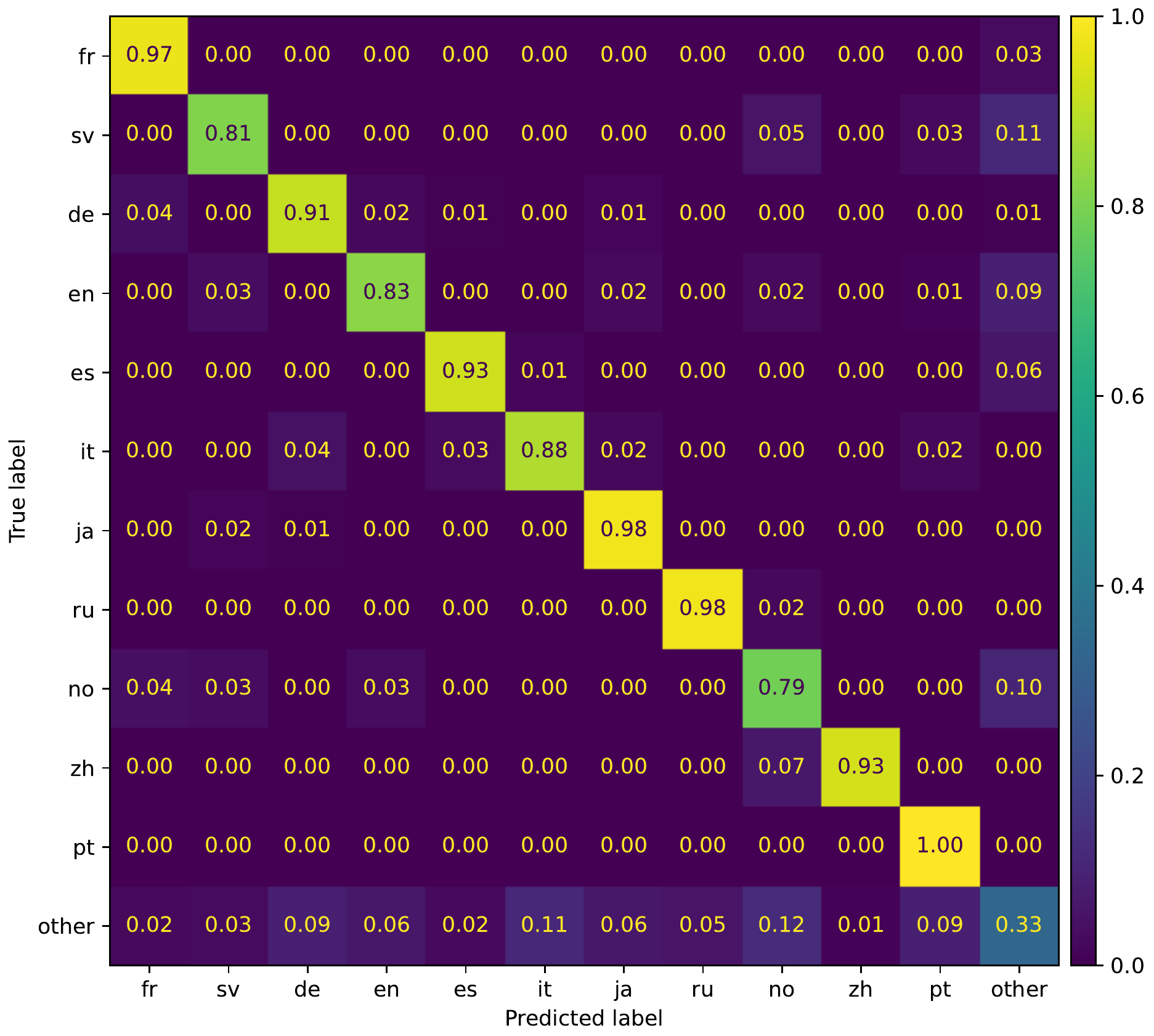}}
  \end{subfigure}
\vspace{-5px}
\caption{Normalized confusion matrices for LECAPAT on VoxLingua11+O for multiclass (left) and multilabel (right).}
\label{fig:confusion}
\vspace{-0.1cm}
\end{figure}

\begin{table}[t]
\begin{center}
\begin{tabular}{ l c c}
 Model & Multiclass & Multilabel \\ 
 \hline
 LECAPAT &  22.56  & \textbf{14.85}\\
 TC-ResNet10 & 29.02 & \textbf{23.93} \\
 TC-ResNet14 & 32.07 & \textbf{31.01} \\
\end{tabular}
\vspace{0.1cm}
\caption{Error rates for our proposed models using the multiclass and multilabel training strategies on VoxLingua11+O.}
\label{tab:multilabel}
\vspace{-0.8cm}
\end{center}
\end{table}

Finally, we examine how our proposed multilabel training strategy for handling non-target languages at inference time (cf.~Section \ref{sub:multilabelclass}) compares to a multiclass setup which explicitly models them via an additional ``Other'' class. 
For this experiment, we train both multiclass and multilabel models with the full VoxLingua107 training set.
The results for our proposed models on VoxLingua11+O using both strategies are reported in Table~\ref{tab:multilabel}. 
We note the error rates are generally higher than those for VoxLingua11. This is expected, as the classification problem is more challenging now given the presence of the additional, highly heterogeneous, class ``Other''. It also shows that studies which only evaluate SLR against a known set of target languages are most likely not representative of how models would perform under deployment conditions where non-target languages may be presented to the model.
We see that our multilabel strategy significantly outperforms the multiclass setup for all three models, with the improvement being most notable for our top-performing LECAPAT model. We hypothesize that the multiclass model struggles to learn an embedding space that can model a large number of non-target languages as a single ``Other'' class while keeping it disjoint from the target languages. To explore this, we plot the normalized confusion matrices for LECAPAT trained with each of the two strategies in Figure \ref{fig:confusion}. 
We see the multiclass model (left) is able to better classify non-target languages as ``Other,'' but suffers from notable leakage of target classes into the ``Other'' class.
The multilabel model (right) alleviates this considerably, e.g., improving the accuracy for Spanish, Italian, Russian, and Norwegian by over 10 percentage points each.

\section{Conclusion}
\label{sec:conclusions}

In this paper we addressed the problem of efficient Spoken Language Recognition (SLR) in the presence of non-target languages. To the best of our knowledge this is the first study on efficient SLR, and the first to propose multilabel training to handle non-target languages at inference time. Our experiments show that our top-performing proposed model, LECAPAT, performs almost on par with models that are two OOM larger, and only 6 accuracy points lower than the largest models while being four OOM smaller and almost two OOM faster. Additionally, we show that our proposed multilabel training strategy outperforms the multiclass setting by a considerable margin when non-target languages are present at inference time. We hope this study will stimulate further work on the topic of efficient SLR under real-world conditions.

\bibliographystyle{IEEEtran}
\bibliography{main}

\end{document}